\documentclass[conference]{IEEEtran}
\usepackage{times}

\usepackage[numbers]{natbib}
\usepackage{multicol}
\usepackage{bm}
\usepackage{amsmath}
\usepackage{amsthm}
\usepackage{amssymb}
\usepackage{color}
\usepackage{subcaption}
\usepackage{mathtools}
\usepackage[bookmarks=true]{hyperref}
\usepackage{adjustbox}
\usepackage{array}

\newcolumntype{R}[2]{%
    >{\adjustbox{angle=#1,lap=\width-(#2)}\bgroup}%
    l%
    <{\egroup}%
}

\theoremstyle{definition}

\newtheorem{theorem}{Theorem}

\newtheorem{remark}{Remark}

\begin{document}

\title{Kalman Filtering with Gaussian Processes Measurement Noise}

\author{Vince Kurtz, Hai Lin}



%

\maketitle

\begin{abstract}
Real world measurement noise in applications like robotics is often correlated in time, but we typically assume i.i.d. Gaussian noise for filtering. We propose general Gaussian Processes as a non-parametric model for correlated measurement noise that is flexible enough to accurately reflect time-correlated measurement noise, yet simple enough to enable efficient computation. We show that this model accurately reflects the measurement noise resulting from vision-based Simultaneous Localization and Mapping (SLAM), and argue that it provides a flexible means of modeling measurement noise for a wide variety of sensor systems and perception algorithms. We then extend existing results for Kalman filtering with autoregressive processes to more general Gaussian Processes, and demonstrate the improved performance of our approach.
\end{abstract}

\IEEEpeerreviewmaketitle

\section{Introduction}

\subsection{Motivation}

Robotic systems often rely on advanced perception algorithms and complex sensor suites. Such perception algorithms provide measurements of position, velocity, and other relevant quantities, but the associated measurement noise is often correlated in time. Traditional Kalman filtering assumes white i.i.d. Gaussian measurement noise, and as such is not optimal for these applications. 

To address this gap, we propose using general (non-white) Gaussian Processes (GPs) as a non-parametric noise model that can capture the correlation present in these perception systems. As a Bayesian non-parametric model, GPs can capture a wide variety of qualitative behaviors. They are direct generalizations of the commonly used independent Gaussian and AR(p) noise models. At the same time, they retain many desirable computational properties, as they are based on multivariate Gaussian distributions, for which many mature mathematical and computational techniques exist. 

\subsection{Related Work}

Most existing work on filtering with correlated measurement noise has focused on using autoregressive (AR) noise models. \citet{bryson1968estimation} provide the earliest work on optimal filtering with an AR(1) noise model, taking a state augmentation approach to transform the problem into one of Kalman filtering with white noise. More recently, \citet{petovello2009consideration} provided an alternative approach for AR(1) noise models using a ``time-differencing'' approach.

Much recent research has focused on extending these approaches (state augmentation and time-differencing) to more general noise models. AR(p) and autoregressive moving average (ARMA) models were studied in \cite{geist2011kalman}, \cite{liu2015optimal} and \cite{jiang2010globally}. \cite{chang2014kalman} proved the theoretical equivalence of the time-differencing and state augmentation approaches. They note that accounting for correlation in the noise both improves the resulting estimate and results in more conservative error convariances, a measure of uncertainty in the estimate. We find a similar effect in this work for the case of GP measurement noise. 

A common approach to filtering with observations perturbed by correlated noise is to use a pre-whitening filter. In particular, given measurments $\mathbf{z} = \mathbf{x} + \mathbf{v}$, where $\mathbf{v} \sim \mathcal{N}(0,\Sigma)$ and $\Sigma$ has off-diagonal elements, we might apply a prewhitening transformation $U$ such that $U\mathbf{v} \sim \mathcal{N}(0,U\Sigma U^T)$ and $U\Sigma U^T$ is diagonal. 

To use this approach, however, we need to have access to all of the measurements $\mathbf{z}$ first. In robotics applications, these measurements a stream of data coming in in real time. At each step, we need to update our state estimate using the latest measurement. The prewhitening approach would require re-computing the transformation $U$ at each timestep as new measurements arrive and $\mathbf{z}$ increases in size, which might be computationally prohibitive. 

Instead, we take our inspiration from \cite{jiang2010globally}, which provides a recursive algorithm for optimal filtering for AR(p) measurement noise, which is derived directly as the minimum mean squared error (MMSE) linear filter. This formulation allows us to analytically derive an optimal filter for the GP noise case without resorting to state augmentation or prewhitening. 

In the control literature, GPs have been used to model the state transition and measurement functions, and new optimal filtering techniques have been developed for such models \cite{ko2009gp,deisenroth2012robust}. But to the best of our knowledge, this is the first work to propose using a general GP to model the measurement noise itself. 

\section{Preliminaries}

\subsection{Kalman Filter}

Kalman filtering is concerned with estimating the state of the linear state-space model
\begin{equation}\label{eq:kalman_sys}
    \begin{gathered}
    x_{t+1} = F_tx_t + w_t \\
    z_t = H_tx_t + v_t
    \end{gathered}
\end{equation}
where $x_t \in \mathbb{R}^n$ is the underlying state and $z_t \in \mathbb{R}^m$ are noisy observations. $w_t \sim \mathcal{N}(0,W_t)$ and $v_t \sim \mathcal{N}(0,V_t)$ represent the process and measurement noise respectively. The classical Kalman filter assumes that $w_t$ and $v_t$ are white and mutually uncorrelated. That is, 
\begin{gather*}
    E[v_t,w_{t^\prime}] = 0 ~~~~ \forall ~ t,t^\prime \\
    E[w_t,w_{t^\prime}] = W_t\delta(t,t^\prime) \\
    E[v_t,v_{t^\prime}] = V_t\delta(t,t^\prime)
\end{gather*}
where $\delta(i,j)$ is the Kronecker delta. 

Given observation $z_k$, the updated state estimate $\hat{x}_k$ and estimate covariance $P_k \in \mathbb{R}^{n\times n}$ are calculated in two steps as follows:

First, system (\ref{eq:kalman_sys}) is used to update the estimate and propagate the uncertainty in the estimate.
\begin{gather*}
    \hat{x}_k^- = F_{k-1}\hat{x}_{k-1} \\
    P_k^- = F_{k-1}P_{k-1}F_{k-1}^T + W_k
\end{gather*}
Then, the measurement $z_k$ is used to refine this estimate and the associated covariance.
\begin{gather*}
    K_k = P_k^-H_k^T[V_k + H_kP_k^-H_k^T]^{-1} \\
    \hat{x}_{k} = \hat{x}_k^-+K_k(z_k-H_k\hat{x}_k^-) \\
    P_k = (I - K_kH_k)P_k^-
\end{gather*}

\subsection{Gaussian Process}

A Gaussian Process (GP) is a random process where every subset is a multivariate Gaussian random vector. GPs can be uniquely specified with a mean function $\mu(t)$ and a covariance function $k(t,t')$ as follows:

\begin{equation}
    \mathbf{x} \sim \mathcal{GP}(\mu(t), k(t,t'))
\end{equation}
\begin{equation}
    \begin{bmatrix}
        \mathbf{x}_{t_1} \\
        \mathbf{x}_{t_2} \\
        \vdots \\
        \mathbf{x}_{t_N}
    \end{bmatrix} \sim
    \mathcal{N}(\bm{\mu}, \mathbf{K})
\end{equation}
where $[\bm{\mu}]_{t_i}=\mu(t_i)$ and $\mathbf{K}$ is the Gram matrix defined such that $[\mathbf{K}]_{i,j}=k(t_i,t_j)$. In most cases, the mean function is assumed to be zero. 

Gaussian Processes are an example of a \textit{Bayesian Non-parametric} model, in the sense that the parameters $\bm{\mu}$ and $\mathbf{K}$ are not specified directly, but rather determined by the hyperparameters of $\mu(t)$ and $k(t,t^\prime)$. Such hyperparameters can be determined from data using Bayesian techniques \cite{gelman2013bayesian}. 

For example, one commonly used Kernel function is the squared exponential, or RBF, kernel
\begin{equation}
    k(t,t') = exp(\frac{\|t-t'\|^2}{2l^2})
\end{equation}
The RBF kernel is characterized by a lengthscale $l$, which (intuitively) regulates how close the indices $(t,t^\prime)$ must be for $\mathbf{x}_t$ and $\mathbf{x}_{t^\prime}$ to be closely correlated. GPs with RBF kernels give rise to functions that are infinitely differentiable. Other kernel functions, such as the popular Mat\'ern class of kernels, describe less smooth functions, while others yet model periodic correlations \cite{rasmussen2004gaussian}. 

For a more comprehensive view of Gaussian processes, kernel functions, and their applications, we refer the interested reader to \cite{roberts2013gaussian,rasmussen2004gaussian}. 

\section{Gaussian Processes as Noise Model}

A growing body of work in the control systems community has focused on Gaussian processes for system identification of stochastic systems. In this section, we propose Gaussian Processes as a flexible model of correlated measurement noise. 

Specifically, we consider a modification of System (\ref{eq:kalman_sys}) such that rather than being uncorrelated in time, $v_t$ is drawn from a zero-mean Gaussian Process:
\begin{equation}\label{eq:gp_noise_model}
    v_t \sim \mathcal{GP}(0,k(t,t^\prime)).
\end{equation}


This formulation has a good deal of expressive power, as GPs can model a wide range of qualitative behaviors \cite{rasmussen2004gaussian,roberts2013gaussian}. This flexibility suggests that GPs can accurately model noise for a wide variety of sensor configurations and perception algorithms. Furthermore, the GP noise model generalizes both the common independent Gaussian noise model and the popular AR(p) model \cite{davis2014gaussian}. At the same time, GPs are simple enough to maintain useful computational properties, since every subset is a multivariate Gaussian. 

\subsection{Determining Model Hyperparameters}

Much as a traditional independent Gaussian noise model requires determining a value of the covariance parameter $V_t$, using GPs as a noise model requires the specification of a kernel function, either from prior knowledge or from data. In this section, we demonstrate how Bayesian techniques can be used to determine a kernel's hyperparameters from data. 

GPs are an example of a \textit{Bayesian Non-parametric} model \cite{roberts2013gaussian}. As such, we do not infer any parameters directly, but rather reason about hyperparameters $\phi$. These hyperparameters characterize the kernel function, and can give rise to a wide variety of qualitative behaviors. 

To obtain posterior inference on the hyperparameters $\phi$, we must consider the likelihood $f(\mathbf{v} | \phi)$ where $\mathbf{v} = [v_0^T,v_1^T,\dots]^T$. The log likelihood is given by
\begin{equation}\label{eq:lml}
    \log f(\mathbf{v} | \phi) = -\frac{1}{2}\mathbf{v}^T K_\mathbf{v}^{-1} \mathbf{v} - \frac{1}{2} \log |K_\mathbf{v}| - \frac{n}{2}\log 2\pi
\end{equation}
where
$$
K_\mathbf{v} = 
\begin{bmatrix}
    k(t_1,t_1) & k(t_2,t_1) & \ldots                        \\
    \vdots                       & \ddots                       &                               \\
    k(t_n,t_1) &                              & k(t_n,t_n)   \\
\end{bmatrix}
$$
is specified according to the hyperparameters $\phi$ \cite{rasmussen2004gaussian}.

\begin{figure*}
    \begin{subfigure}{0.5\textwidth}
        \centering
        \includegraphics[width=\linewidth]{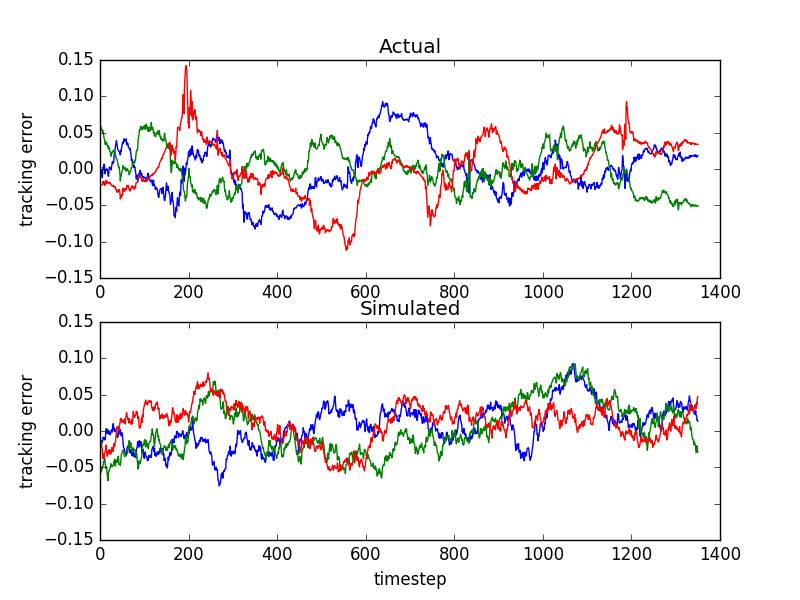}
        \caption{Best-fit (ML-II) Gaussian Process noise model with exponential kernel.}
        \label{fig:rgbdslam_example_gp}
    \end{subfigure}
    \begin{subfigure}{0.5\textwidth}
        \centering
        \includegraphics[width=\linewidth]{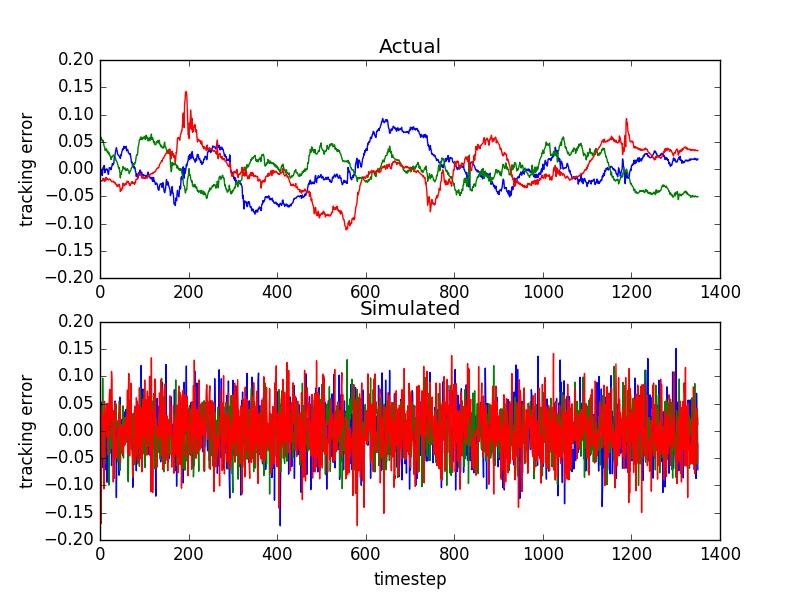}
        \caption{Best-fit (Maximum a-postiori) i.i.d. Gaussian noise model.}
        \label{fig:rgbdslam_example_iid}
    \end{subfigure}
    \caption{Actual measurement noise (x,y,z) from RGBDSLAM \cite{endres20143} (top) compared to samples from a corresponding noise model. The Gaussian Process model captures autocorrelation in the noise, leading to error dynamics that are qualitatively similar to the actual error, while the typically i.i.d. Gaussian model does not account this autocorrelation.}
    \label{fig:rgbdslam_example}
\end{figure*}

Given this log likelihood function, we can use Bayes rule to determine the posterior disribution of the hyperparameters $\phi$:
\begin{equation}
    f(\phi|\mathbf{v}) \propto f(\mathbf{v}|\phi)f(\phi).
\end{equation}

\subsection{Case Study: RGBDSLAM}

In this case study, rather than finding the exact posterior distribution of the hyperparameters, we simply maximize the likelihood given in (\ref{eq:lml}), a technique known as ML-II \cite{rasmussen2004gaussian}. We model the noise $v_t$ as being drawn from a zero-mean stationary and isotropic GP. Specifically, we assume that the measurement noise on each axis, (x,y,z), is drawn from the same underlying process with an exponential kernel,
\begin{equation}\label{eq:exp}
    k(r) = \sigma^2 exp(-\frac{r}{l}).
\end{equation}

The exponential kernel is a special case of the Matern kernel with degrees of freedom $\nu=1/2$. Functions drawn from a GP with an exponential kernel are once differentiable and correspond to the Ornstein-Ulenback (OU) process. The OU process describes a random walk which tends to return to the mean: intuitively reasonable behavior when considering localization error. 

RGBD-SLAM \cite{endres20143} is a standard SLAM algorithm that takes as input measurements from a RGB-D camera and outputs an estimated position and orientation. As a case study, we run RGBDSLAM on the ``freiburg1\_room'' scenario from the TUM-vision dataset \cite{sturm12iros}. ML-II yields hyperparameter values of $\sigma^2=1.2\times10^{-3}$ and $l=135$. The results ($N=100$) are shown in Figure \ref{fig:rgbdslam_example_gp}. 

\begin{figure*}
    \begin{subfigure}{0.5\textwidth}
        \centering
        \includegraphics[width=0.9\textwidth]{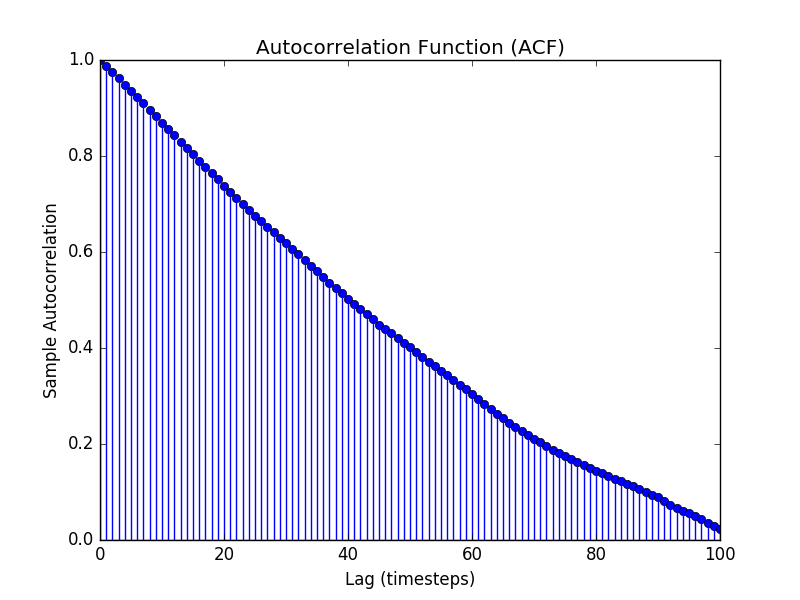}
        \caption{Autocorrelation function plot for RGBDSLAM data}
        \label{fig:acf_rgbdslam}
    \end{subfigure}
    \begin{subfigure}{0.5\textwidth}
        \centering
        \includegraphics[width=0.9\textwidth]{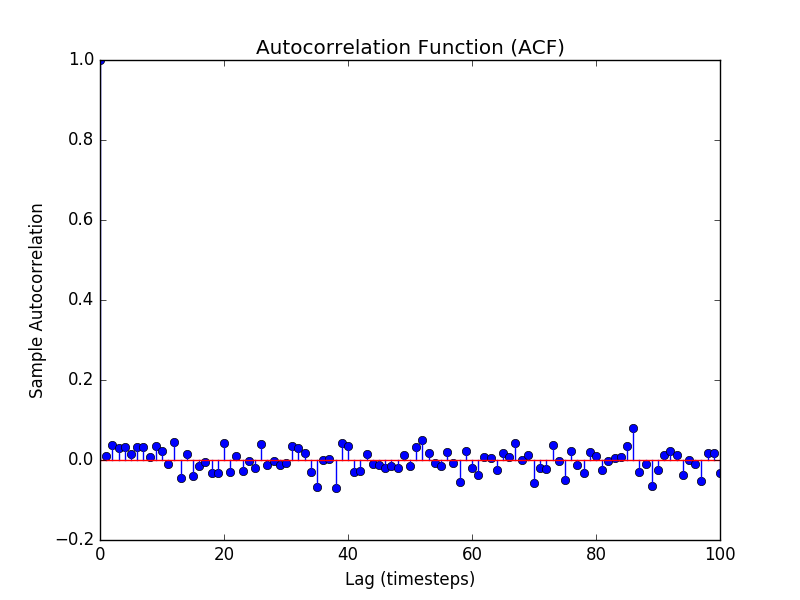}
        \caption{Autocorrelation function plot for i.i.d. Gaussian data}
        \label{fig:acf_gaussian}
    \end{subfigure}
    \caption{Actual data from RGBDSLAM shows significant sample autocorrelation, while (simulated) i.i.d. Gaussian data shows close to none.}
    \label{fig:acf}
\end{figure*}

We can quantify the auto-correlation present in the actual tracking error with the use of an autocorrelation function (ACF) plot, which essentially displays the sample autocorrelation for different timesteps (lags). An ACF plot for the y-axis error from RGBDSLAM and an ACF plot for i.i.d. Gaussian noise are shown in Figure \ref{fig:acf}.

\section{Kalman Filtering with GP Measurement Noise}

Given a complex perception system (such as RGBDSLAM or AprilTags) that provides noisy observations, we may wish to make an optimal estimate of the robots's state. A classical approach at this point would be to use a Kalman filter; however, as demonstrated above, the noisy observations produced by today's advanced perception algorithms contain significant autocorrelation. This noise is thus better modeled by Gaussian Processes than by the i.i.d. Gaussian noise assumed by the Kalman filter. In this section, we derive a linear unbiased estimator of the state $x_t$ given observations $Z_t = [z_1~z_2\dots z_t]^T$ that is optimal in the mean-square-error (MSE) sense. 

We take as our starting point the following standard intermediate result (stated as found in \cite{jiang2010globally}) from proving the optimality of the Kalman filter:

\begin{theorem}
    Regardless of time correlation in $v_t$, the optimal estimate of $x_t$, $\hat{x}_t$, is given by
    \begin{align}
        & \hat{x}_t^- = F_{t-1}\hat{x}_{t-1}  \\
        & P_t^- = F_{t-1}P_{t-1}F_{t-1}^T + W_t
    \end{align}
    \begin{align}
        & \hat{x}_t = \hat{x}_t^- + K_t(z_t - \hat{z}_t)\label{eq:optimal_est}\\
        & P_t = P_t^- - K_tJ_t^T\label{eq:optimal_var}
    \end{align}
    where    
    \begin{align*}
        & \hat{z}_t = E[z_t] + Cov(z_t,Z_{t-1})Var(Z_t)^{-1}(Z_{t-1}-E[Z_{t-1}]), \\
        & L_t = Var(z_t) + Cov(z_t,Z_{t-1})Var(Z_t)^{-1}Cov(Z_{t-1},z_t), \\
        & J_t = P_t^-H_t^T, \\ 
        & K_t = J_tL_t^{-1}.
    \end{align*}
\end{theorem}

Note that in the case of i.i.d. measurement noise $Cov(z_t,Z_t) = 0$, $\hat{z}_t = H_t\hat{x}_t^-$, and we recover the traditional Kalman filter.

For the case of Gaussian Process noise, we proceed as follows. Note that  
\begin{equation}
    E[Z_t] = E[\begin{bmatrix}z_t \\ z_{t-1} \\ \vdots \end{bmatrix}] = \begin{bmatrix}H_t\hat{x}_t^- \\ H_{t-1}\hat{x}_{t-1}^- \\ \vdots \end{bmatrix},
\end{equation}
\begin{equation}
    Cov(z_t,Z_{t-1}) = Cov(Z_{t-1},z_t)^T = [k(t,t+1)~k(t,t+2)\dots],
\end{equation}
and
\begin{multline}
    Var(Z_t) = Var(\begin{bmatrix}z_t \\ z_{t-1} \\ \vdots \end{bmatrix}) = \\
    \begin{bmatrix}
        H_tP_t^-H_t^T & 0                         & \dots \\
        0             & H_{t-1}P_{t-1}^-H_{t-1}^T &       \\
        \vdots        &                           & \ddots
    \end{bmatrix}+ \\
    \begin{bmatrix}
        k(t,t)   & k(t,t+1)   & \dots \\
        k(t+1,t) & k(t+1,t+1) &       \\
        \vdots   &            & \ddots
    \end{bmatrix}\label{eq:var_Z}.
\end{multline}

With this, we can compute $L_t, J_t$, and $\hat{z}_t$. The optimal estimate of the state $x_t$ is thus given by (\ref{eq:optimal_est}), and its associated error covariance by (\ref{eq:optimal_var}).

\subsection{Practical Kalman Filtering}

The procedure above gives the optimal estimate, which we now denote $\hat{x}^*_t$, but this estimate depends on the whole history of observations $Z_t$. Furthermore, this approach requires inverting $Var(Z_t)$, a $t\times t$ matrix which grows at every timestep. To avoid this increasing complexity, we propose to consider only a truncated version of the measurement history, $\hat{Z}_t = [z_{t-N+1}~z_{t-N+2}\dots z_{t}]$. 

By considering only this ``window'' of $N$ measurements, we can still compute an estimate $\hat{x}_t$ using (\ref{eq:optimal_est},\ref{eq:optimal_var}) and replacing $Z_t$ with $\hat{Z}_t$. But this time, the size of $Var(\hat{Z}_t)$ is fixed at $(N\times N)$. 

Furthermore, using a fixed size window allows us to take advantage of the structure of (\ref{eq:var_Z}) to more efficiently compute an estimate. Specifically, given a kernel function $k(\cdot,\cdot)$ and a window length $N$ we can precompute
\begin{equation*}
    A^{-1} = \begin{bmatrix}
        k(t,t)   & k(t,t+1)   & \dots \\
        k(t+1,t) & k(t+1,t+1) &       \\
        \vdots   &            & \ddots
    \end{bmatrix}^{-1}.
\end{equation*} Note that we can decompose the other term in $Var(\hat{Z})$ as
\begin{multline*}
    \begin{bmatrix}
        H_tP_t^-H_t^T & 0                         & \dots \\
        0             & H_{t-1}P_{t-1}^-H_{t-1}^T &       \\
        \vdots        &                           & \ddots
    \end{bmatrix} = UV = \\
    \begin{bmatrix}
        H_tP_t^- \\ H_{t-1}P_{t-1}^- \\ \vdots
    \end{bmatrix}
    \begin{bmatrix}
        H_t^T & H_{t-1}^T & \dots
    \end{bmatrix}.
\end{multline*}

\begin{figure*}
    \begin{subfigure}{0.5\textwidth}
        \centering
        \includegraphics[width=0.9\textwidth]{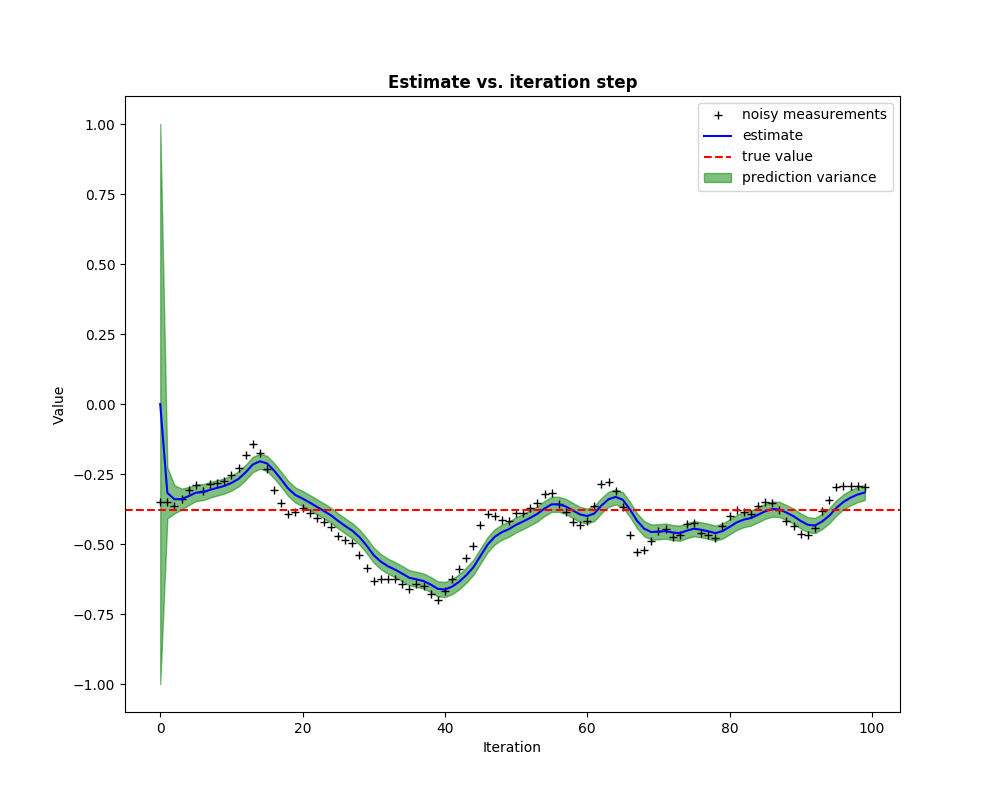}
        \caption{Kalman Filter}
    \end{subfigure}
    \begin{subfigure}{0.5\textwidth}
        \centering
        \includegraphics[width=0.9\textwidth]{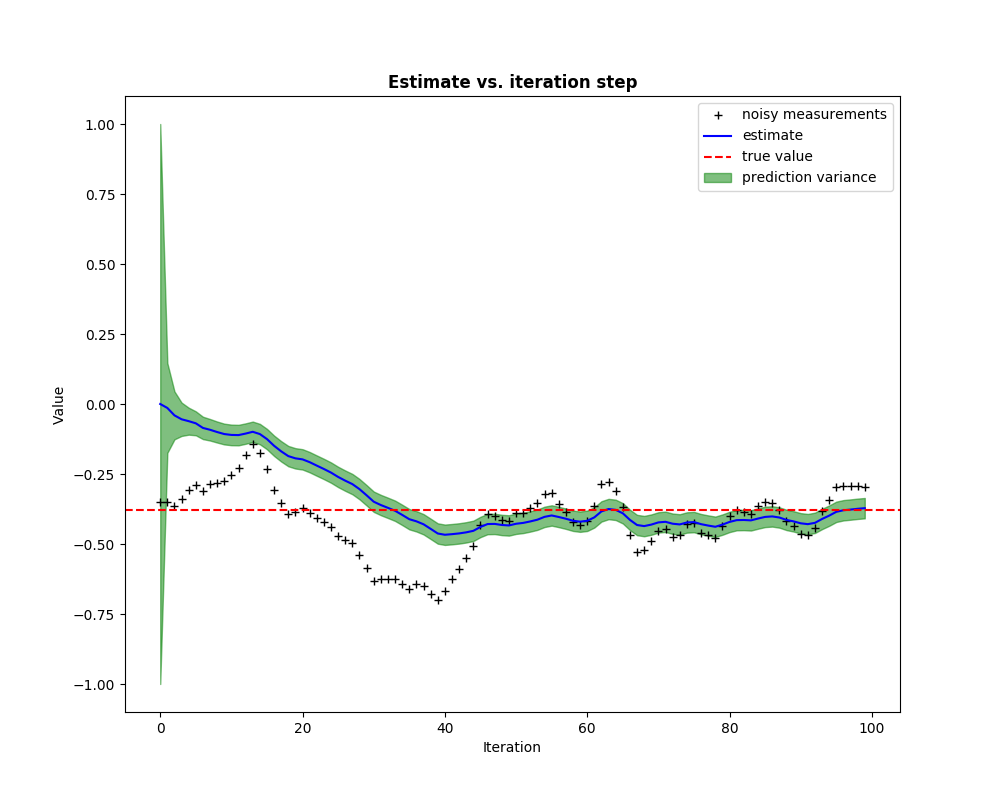}
        \caption{Our filter with full history}
    \end{subfigure}
    \begin{subfigure}{0.5\textwidth}
        \centering
        \includegraphics[width=0.9\textwidth]{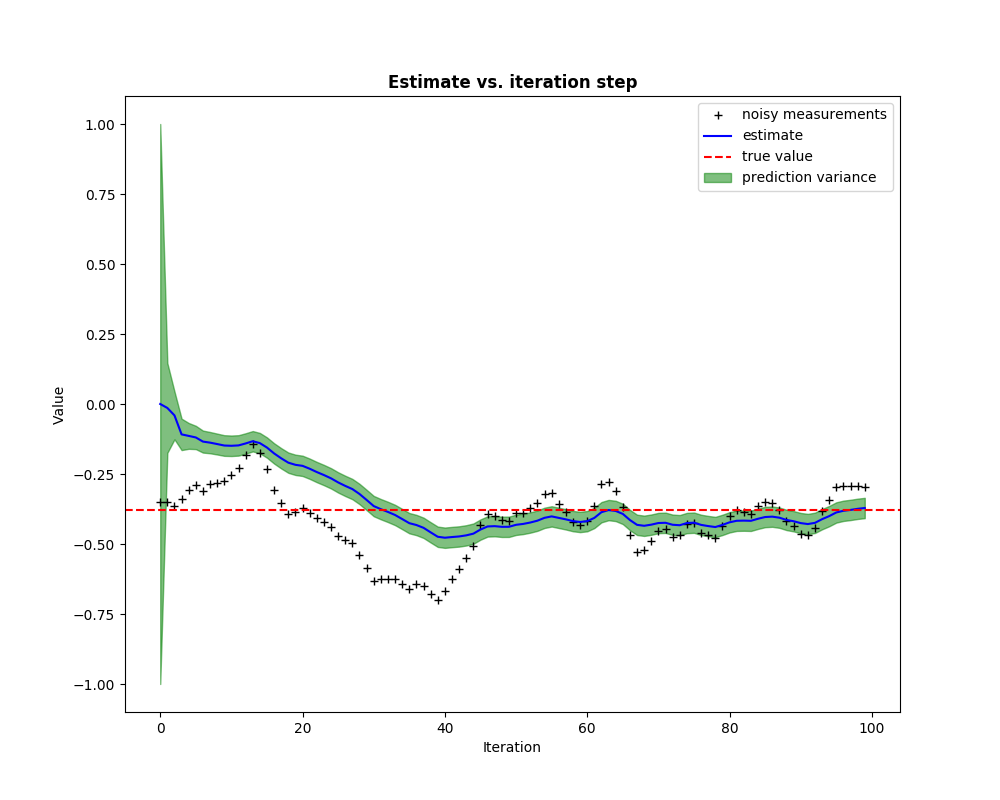}
        \caption{Our filter with N=2}
    \end{subfigure}
    \begin{subfigure}{0.5\textwidth}
        \centering
        \includegraphics[width=0.9\textwidth]{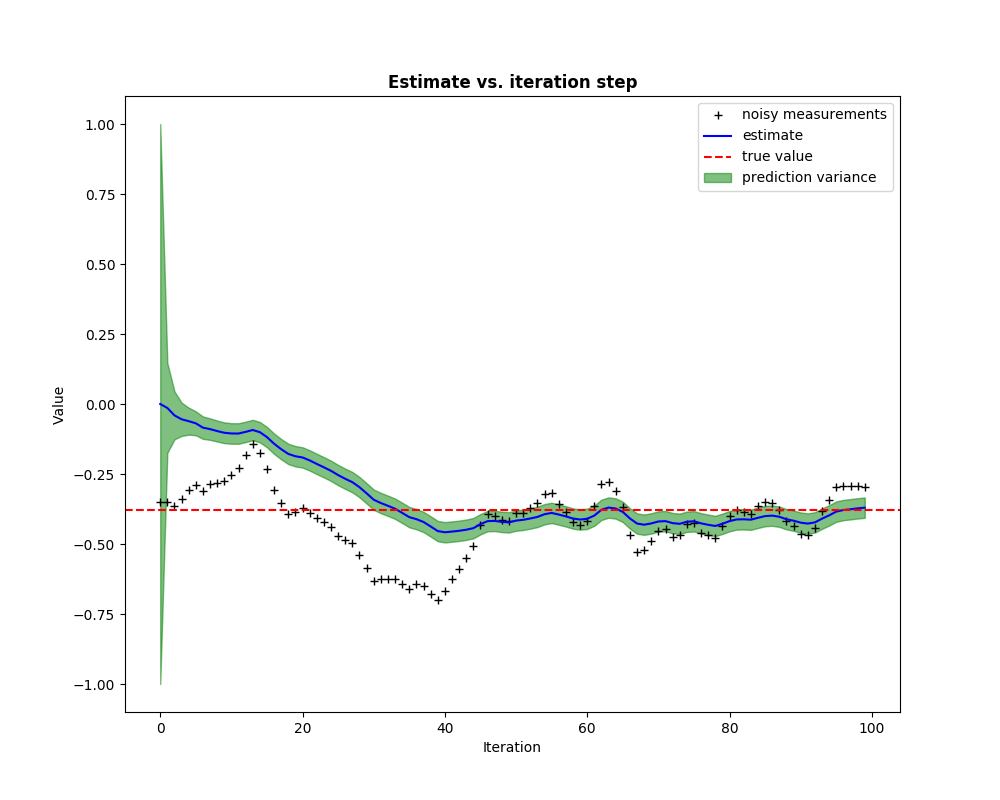}
        \caption{Our filter with N=5}
    \end{subfigure}
    \caption{Comparison of filtering paradigms for estimating a scalar value via noisy observations drawn from a Gaussian Process (Matern kernel, lengthscale 5).}
    \label{fig:matern_example}
\end{figure*}

We can then use the Sherman-Morrison-Woodbury matrix inversion lemma to compute $Var(\hat{Z})^{-1}$ as follows:

\begin{multline*}
    Var(\hat{Z})^{-1} = (A+UCV)^{-1} = \\ A^{-1} - A^{-1}U(I_{m \times m} + VA^{-1}U)^{-1}VA^{-1}.
\end{multline*}

\begin{remark}
    $Var(\hat{Z})$ is an $Nn\times Nn$ matrix, and thus computing $Var(\hat{Z})^{-1}$ directly is an $O((Nn)^3)$ process. But if we precompute $A^{-1}$ and use the matrix inversion lemma as described above, we only need to compute the inverse of $I + VA^{-1}U$, an $m \times m$ matrix, at each timestep. Using this technique renders the complexity of calculating $\hat{x}_t$ linear in the size of the history window, i.e. $O(N)$. In practice, this means that we can choose quite a large $N$ without worrying about computational speed. 
\end{remark}

From an engineering perspective, there is tradeoff that regulates the choice of the size of the window $N$. Choosing a larger $N$ generally results in better performance, since more observations are taken into account. At the same time, a larger $N$ increases the complexity of computing the estimate $\hat{x}_t$, though as stated above, this complexity is only $O(N)$. 

One option for choosing $N$ would be to choose the largest possible $N$ that allows for realtime computation of the state estimate. A more insightful way might be to note that most kernel functions decay as $t'-t$ increases, leading to behavior like that shown in Figure \ref{fig:acf_rgbdslam}. By setting some correlation threshold, we might choose some minimum coorelation $k_{min}$ and choose $N$ such that we neglect only those observations so far in the past that their correlation is below the threshold, i.e.,
\begin{align*}
    \min &~ N \\
    \text{s.t.} &~ k(t,t+l) < k_{min} ~~~\forall~l\geq N.
\end{align*}

An alternative heuristic would be to start the estimation by using all available estimates $Z_t$, and determine $N$ on-the-fly based on the error covariance $P_t$. When $P_t$ falls below a certain threshold, we set $N$ and start moving the window of observations $\hat{Z}_t$. Since the error covariance generally decreases over time, this heuristic allows us to choose of $N$ in terms of the filtering error. 

\section{Example}

As a simple example, consider estimating a scalar value from observations perturbed by noise drawn from a GP (Matern Kernel, lengthscale of 5), as shown in Figure \ref{fig:matern_example}.

Using the Kalman filter (or equivalently, our filter with $N=1$) results in an overly optimistic estimate that follows the observations too closely: the assumption of uncorrelated noise leads to overaggressive following of the observations. Additionally, the error covariance $P_t$ (represented by the green shaded region) seems to be too small. 

Using our filter with the full full history $Z_t$ causes the estimate to follow the observations less closely and converge more steadily to the true value, but the computation is significantly slower, since this requires inverting matrices as large as $(100\times 100)$.

With a truncated history $\hat{Z}_t$, using as few as $N=2$ steps already shows a marked improvement in performance. The resulting estimate with $N=5$ is virtually indistinguishable from the optimal estimate. This suggests that the GP lengthscale hyperparameter may be yet another useful heuristic for choosing how many steps of history to account for. 

\section{Conclusion}

We presented general Gaussian Processes as a flexible model of measurement noise. We showed that advanced sensor systems and perception algorithms such as RGBDSLAM produce measurement errors that are highly correlated in time, and well modeled by GPs. We derived an optimal Kalman filter for the case of general Gaussian Process noise, and demonstrated in simulation that accounting for autocorrelations in measurement noise via a GP model leads to superior performance. 

\bibliographystyle{plainnat}
\bibliography{references}

\end{document}